# A note on communicating between information systems based on including degrees

Ping Zhu[a,b*] and Qiaoyan Wen[b†]

[a]*School of Science, Beijing University of Posts and Telecommunications, Beijing 100876, China*; [b]*State Key Laboratory of Networking and Switching, Beijing University of Posts and Telecommunications, Beijing 100876, China*

In order to study the communication between information systems, Gong and Xiao [Z. Gong and Z. Xiao, Communicating between information systems based on including degrees, International Journal of General Systems 39 (2010) 189–206] proposed the concept of general relation mappings based on including degrees. Some properties and the extension for fuzzy information systems of the general relation mappings have been investigated there. In this paper, we point out by counterexamples that several assertions (Lemma 3.1, Lemma 3.2, Theorem 4.1, and Theorem 4.3) in the aforementioned work are not true in general.

## 1. Preliminaries

In this section, we recall some basic notions of Pawlak's rough set theory (Pawlak 1982, 1991) and the concept of general relation mappings based on including degrees due to Gong and Xiao (2010).

Let $U$ be a finite and nonempty universal set. We write $\Re(U)$ for the set of all equivalence relations on $U$. For any $R \in \Re(U)$, denote by $U/R$ the set of all equivalence classes induced by $R$. For any $x \in U$, we write $[x]_R$ for the equivalence class induced by $R$ that contains $x$. Formally, $[x]_R = \{y \in U \mid (x,y) \in R\}$. For any $X \subseteq U$, one can characterize $X$ by a pair of lower and upper approximations. The *lower approximation* $\underline{\mathrm{apr}}_R X$ and *upper approximation* $\overline{\mathrm{apr}}_R X$ of $X$ are defined as follows:

$$\underline{\mathrm{apr}}_R X = \cup \{C \in U/R \mid C \subseteq X\},$$
$$\overline{\mathrm{apr}}_R X = \cup \{C \in U/R \mid C \cap X \neq \emptyset\}.$$

To state the notion of general relation mappings, it is convenient to recall the following concept of including degrees from (Gong and Xiao 2010).

**Definition 1.1:** (Gong and Xiao 2010, Definition 2.3) Let $U$ be a finite set and $\mathcal{P}$ the power set of $U$. The *including degree* on $\mathcal{P}$ is defined as

$$D(F/E) = |E \cap F|/|E|,$$

---

[*]Corresponding author. Email: pzhubupt@gmail.com

[†]Email: wqy@bupt.edu.cn

where $E, F \in \mathcal{P}$ and "$|S|$" denotes the cardinality of a set $S$.

**Definition 1.2:** (cf. Gong and Xiao 2010, Definition 3.1) Let $U$ and $V$ be finite universes, and $f : U \longrightarrow V$ a surjective mapping. The *general relation mapping* induced by $f$, denoted by the same notation $f$, is a mapping from $\mathfrak{R}(U)$ to $\mathfrak{R}(V)$ defined by

$$f(R) = \{(f(x), f(y)) \mid (x, y) \in R \wedge D([x]_R/[x]_f) = D([y]_R/[y]_f)\}$$

for all $R \in \mathfrak{R}(U)$, where $[x]_f = \{y \in U \mid f(y) = f(x)\}$.

**Remark 1:** In Definition 3.1 and some results such as Theorem 3.1 and Lemma 3.1 in (Gong and Xiao 2010), the mapping $f$ is not required to be surjective. In fact, if $f$ is not surjective, then there exists $v \in V - f(U)$ such that $(v, v) \notin f(R)$, which implies that $f(R)$ is not an equivalence relation on $V$. Therefore, it is necessary to require that $f$ is surjective.

## 2. Counterexamples

We begin this section with a result given in (Gong and Xiao 2010).

**Lemma 2.1:** (Gong and Xiao 2010, Lemma 3.1) Let $U$ and $V$ be finite universes, and $f : U \longrightarrow V$ a surjective mapping. For any $R_1, R_2 \in \mathfrak{R}(U)$, we have the following:

(1) $R_1 \subseteq R_2$ if and only if $f(R_1) \subseteq f(R_2)$.
(2) $f(R_1 \cap R_2) \subseteq f(R_1) \cap f(R_2)$; the equality holds if $[x]_f \subseteq [x]_{R_i}$, $i = 1, 2$.
(3) $f(R_1 \cup R_2) \supseteq f(R_1) \cup f(R_2)$; the equality holds if $[x]_f \subseteq [x]_{R_i}$, $i = 1, 2$.

The following example shows that some assertions in Lemma 2.1 above are incorrect.

**Example 2.2** Set $U = \{1, 2, \ldots, 6\}$ and $V = \{a, b\}$. Take $R_1 = \{\{1\}, \{2\}, \{3\}, \{4, 5, 6\}\}$ and $R_2 = \{\{3\}, \{1, 2, 4, 5, 6\}\}$, where each element of $R_i$ stands for an equivalence class induced by $R_i$. Define $f : U \longrightarrow V$ as follows:

$$f(1) = f(2) = f(5) = f(6) = a;$$
$$f(3) = f(4) = b.$$

Clearly, $R_1 \subseteq R_2$. However, by a direct computation, we can readily obtain that

$$f(R_1) = \{(a, a), (b, b), (a, b), (b, a)\};$$
$$f(R_2) = \{(a, a), (b, b)\}.$$

As a result, $f(R_2) \subseteq f(R_1)$.

We thus see from the above example that $R_1 \subseteq R_2$ does not imply $f(R_1) \subseteq f(R_2)$. At the same time, we have that $f(R_2) \subseteq f(R_1)$, but $R_2 \not\subseteq R_1$. Consequently, $f(R_1) \subseteq f(R_2)$ does not imply $R_1 \subseteq R_2$ either. Therefore, the assertion (1) in Lemma 2.1 above is incorrect.

In this example, we also see that $f(R_1 \cap R_2) = f(R_1) \supsetneq f(R_2) = f(R_1) \cap f(R_2)$. Hence, the inclusion $f(R_1 \cap R_2) \subseteq f(R_1) \cap f(R_2)$ in the assertion (2) does not hold in general.

In the same example, we also find that $f(R_1 \cup R_2) = f(R_2) \subsetneq f(R_1) = f(R_1) \cup f(R_2)$, which means that the inclusion $f(R_1 \cup R_2) \supseteq f(R_1) \cup f(R_2)$ in the assertion

(3) does not hold in general. More importantly, it should be pointed out that the union of equivalence relations $R_1$ and $R_2$ may not be an equivalence relation, and in this case, $f(R_1 \cup R_2)$ makes no sense.

The following lemma was proved and used in (Gong and Xiao 2010).

**Lemma 2.3:** (Gong and Xiao 2010, Lemma 3.2) Let $U$ and $V$ be finite universes, and $f : U \longrightarrow V$ a surjective mapping. For any $R_1, R_2 \in \mathfrak{R}(U)$, if $[x]_f \subseteq [x]_{R_i}$, $i = 1, 2$, then $f(R_1) - f(R_2) = f(R_1 - R_2)$.

Clearly, $R_1 - R_2$ is not an equivalence relation, and thus $f(R_1 - R_2)$ in the above lemma makes no sense.

Let us consider two theorems in (Gong and Xiao 2010).

**Theorem 2.4:** (Gong and Xiao 2010, Theorem 4.1) Let $U$ and $V$ be finite universes, and $f : U \longrightarrow V$ a surjective mapping. For any $R \in \mathfrak{R}(U)$, we have the following:

(1) $f(\underline{\mathrm{apr}}_R X) \subseteq \underline{\mathrm{apr}}_{f(R)} f(X)$.
(2) $f(\overline{\mathrm{apr}}_R X) \supseteq \overline{\mathrm{apr}}_{f(R)} f(X)$.

**Theorem 2.5:** (Gong and Xiao 2010, Theorem 4.3) Let $U$ and $V$ be finite universes, and $f : U \longrightarrow V$ a surjective mapping. For any $R \in \mathfrak{R}(U)$ and $X \subseteq U$, if $\underline{\mathrm{apr}}_R X = \overline{\mathrm{apr}}_R X = X$, then we have the following:

(1) $f(\underline{\mathrm{apr}}_R X) = \underline{\mathrm{apr}}_{f(R)} f(X) = f(X)$.
(2) $f(\overline{\mathrm{apr}}_R X) = \overline{\mathrm{apr}}_{f(R)} f(X) = f(X)$.

The following example shows the incorrectness of the theorems above.

**Example 2.6** Let $U = \{1, 2, 3, 4\}$ and $V = \{a, b\}$. The mapping $f : U \longrightarrow V$ is defined as follows:

$$f(1) = f(2) = a, \quad f(3) = f(4) = b.$$

Take $R = \{\{1\}, \{2, 3\}, \{4\}\}$ and $X = \{1\}$.

By definition, we have that $\underline{\mathrm{apr}}_R X = \overline{\mathrm{apr}}_R X = X = \{1\}$. Clearly, $f(X) = \{a\}$ and $f(R) = \{(a, a), (b, b), (a, b), (b, a)\}$ by a routine computation. Further, it is easy to get that $\underline{\mathrm{apr}}_{f(R)} f(X) = \emptyset$ and $\overline{\mathrm{apr}}_{f(R)} f(X) = V$.

Therefore, we see that $f(\underline{\mathrm{apr}}_R X) = f(X) = \{a\} \supsetneq \emptyset = \underline{\mathrm{apr}}_{f(R)} f(X)$. It indicates that neither (1) of Theorem 2.4 nor (1) of Theorem 2.5 holds.

In this example, we also obtain that $f(\overline{\mathrm{apr}}_R X) = f(X) = \{a\} \subsetneq V = \overline{\mathrm{apr}}_{f(R)} f(X)$, which means that (2) of Theorem 2.4 and (2) of Theorem 2.5 are not true in general.

By the way, we would like to present another proof of Theorem 4.2 in (Gong and Xiao 2010).

**Theorem 2.7:** (Gong and Xiao 2010, Theorem 4.2) Let $U$ and $V$ be finite universes, and $f : U \longrightarrow V$ a bijective mapping. For any $R \in \mathfrak{R}(U)$, we have the following:

(1) $f(\underline{\mathrm{apr}}_R X) = \underline{\mathrm{apr}}_{f(R)} f(X)$.
(2) $f(\overline{\mathrm{apr}}_R X) = \overline{\mathrm{apr}}_{f(R)} f(X)$.

**Proof :** If $f : U \longrightarrow V$ is a bijective mapping, then we have that $D([x]_R/[x]_f) = 1$ for any $x \in U$. Hence, we get by Definition 1.2 that

$$f(R) = \{(f(x), f(y)) \mid (x,y) \in R \wedge D([x]_R/[x]_f) = D([y]_R/[y]_f)\}$$
$$= \{(f(x), f(y)) \mid (x,y) \in R\}.$$

It follows immediately from Theorem 4.8 in (Wang *et al.* 2008) or Theorem 3.6 in (Zhu and Wen 2010) that both of the assertions are true. $\square$

### Acknowledgements

This work was supported by the National Natural Science Foundation of China under Grants 60821001, 60873191, and 60903152.